\documentclass[preprint,10pt]{elsarticle}
\usepackage[letterpaper,top=2.75cm,bottom=2.75cm,left=2.75cm,right=2.75cm,marginparwidth=2.75cm]{geometry}

\usepackage[T1]{fontenc}
\usepackage[utf8]{inputenc} % (not needed on very new LaTeX, harmless)
\usepackage{microtype}

\usepackage[letterpaper,top=2.75cm,bottom=2.75cm,left=2.75cm,right=2.75cm,marginparwidth=2.75cm]{geometry}

\usepackage{mathtools}  % instead of amsmath; loads amsmath
\usepackage{amsthm,amsfonts,bm}

\usepackage{graphicx,booktabs,multirow,dcolumn,subcaption}
\usepackage{algorithm2e}
\usepackage{listings}

\biboptions{numbers,sort&compress}

\usepackage[backref=page]{hyperref}
\usepackage[nameinlink,capitalize]{cleveref}
\Crefname{equation}{Eq.}{Eqs.}
\crefname{equation}{Eq.}{Eqs.}
\usepackage{xcolor}

\Crefname{equation}{Eq.}{Eqs.}

\usepackage{tikz}
\usepackage{amsmath, amssymb}
\usetikzlibrary{arrows.meta, positioning, calc, decorations.pathreplacing}

\journal{Elsevier}

\begin{document}

\begin{frontmatter}

\title{Multi-fidelity Batch Active Learning for Gaussian Process Classifiers}

%%use optional labels to link authors explicitly to addresses:
\author[1]{Murray Cutforth}
\ead{mcc4@stanford.edu} 
%\cortext[mycorrespondingauthor]{Corresponding author}
\author[2]{Yiming Yang}
\author[1]{Tiffany Fan}
\author[2]{Serge Guillas}
\author[1]{Eric Darve}
\ead{darve@stanford.edu}
\affiliation[1]{address={Mechanical Engineering, Stanford University}}
\affiliation[2]{address={University College London}}

\begin{abstract}
Many science and engineering problems rely on expensive computational simulations, where a multi-fidelity approach can accelerate the exploration of a parameter space. We study efficient allocation of a simulation budget using a Gaussian Process (GP) model in the binary simulation output case.
This paper introduces Bernoulli Parameter Mutual Information (BPMI), a batch active learning algorithm for multi-fidelity GP classifiers. BPMI circumvents the intractability of calculating mutual information in the probability space by employing a first-order Taylor expansion of the link function. 
We evaluate BPMI against several baselines on two synthetic test cases and a complex, real-world application involving the simulation of a laser-ignited rocket combustor. 
In all experiments, BPMI demonstrates superior performance, achieving higher predictive accuracy for a fixed computational budget.

%This linearization creates a computationally efficient approximation that naturally prioritizes queries near the decision boundary, where information gain is maximal, while discounting regions where the classification outcome is already certain. 
\end{abstract}

\begin{keyword}
Active Learning, Bayesian Optimization, Gaussian Process Classification, Multi-fidelity, Uncertainty Quantification
\end{keyword}

\end{frontmatter}

\section{Introduction}
In this paper, we consider the problem of optimally choosing where to run a set of simulations (given some parameter space which we wish to search over) in a bi-fidelity setting, and for binary simulation outputs. This scenario may arise across a variety of science and engineering applications, in which the simulation (or experiment) is prohibitively expensive but can be combined with a larger number of cheaper and less accurate simulations. 

To be more specific, we are studying a laser-ignited methane-oxygen rocket combustor. This is a complex multi-physics problem, where each simulation yields a binary output: ignition or no ignition, which we model as a Bernoulli random variable. Our goal is to obtain the most accurate ignition probability map possible, given two simulation fidelities and a fixed computational budget which is allocated in batches. In some regions of parameter space, the simulation output exhibits stochasticity, making this a probability estimation problem rather than a classification problem.
The batch active learning algorithm developed in this work is generally applicable to any multi-fidelity simulation framework with binary outcomes.

We further limit our scope to Gaussian Process classification models \cite{nickisch2008approximations}, due to their excellent built-in uncertainty quantification abilities. In many engineering applications, it is important to not only predict the quantity of interest, but also to have a robust estimate of the uncertainty on that prediction. We utilize the existing multi-fidelity Gaussian Process classification model of \cite{kennedy2000predicting, costabal2019multi} in this work.

After choosing a suitable bi-fidelity model, the challenge of intelligently exploring the parameter space remains. Given a limited computational budget, one must decide which new simulations to perform (and at which fidelity level) to maximally improve the model's predictive accuracy. This is the problem of active learning (AL), also known as the sequential design of experiments or Bayesian optimization \cite{jones1998efficient}. 
A key component of any AL strategy is the acquisition function, which quantifies the utility of querying a new data point. A particularly powerful class of acquisition functions is based on Mutual Information (MI), which seeks to select points that maximally reduce the uncertainty about a quantity of interest \cite{krause2008near}.

In this work, we develop a novel active learning strategy tailored for bi-fidelity Gaussian Process classification models. Our primary contribution is a new acquisition function, termed Bernoulli Parameter Mutual Information (BPMI), which efficiently approximates the joint mutual information between sets of multi-fidelity Bernoulli parameters in a probability estimation setting.
We demonstrate through numerical experiments that our proposed BPMI strategy outperforms several other mutual information and non-mutual information based active learning strategies. Our numerical experiments focus on estimating the probability map (or classification boundary) in low dimensional (2D) parameter spaces.

\subsection{Related Work}

\paragraph{Multi-fidelity Gaussian Processes} There is a large literature of multi-fidelity Gaussian Process models, starting with the widely-used autoregressive structure of \cite{kennedy2000predicting}. However, the majority of this literature assumes continuous, Gaussian-distributed outputs. Adapting these models to classification settings is non-trivial. The model we use was proposed in \cite{costabal2019multi}.

\paragraph{Active learning} 
Mutual information as an active learning acquisition function was introduced in \cite{krause2008near}. It has been widely used \cite{beck2016sequential}, and is generally optimal if it can be efficiently calculated \cite{takeno2020multi}. There is a significant literature of batched (but single fidelity) active learning methods such as \cite{kirsch2019batchbald}, but a survey of these methods is outside our scope.
To our knowledge, there is only one prior batched, multi-fidelity active learning method in the literature \cite{li2022batch}, which considers real-valued simulation outputs. Our work follows from \cite{li2022batch}, and we develop an efficient algorithm which is specialized to binary outputs and Gaussian Process models.

\section{Method}

We first re-introduce the bi-fidelity Gaussian Process classification model of \cite{costabal2019multi}, before presenting the efficient mutual information-based active learning method which is the focus of this work.

% --- Bi-Fidelity GPC Model Description ---
% Recommended packages: \usepackage{amsmath, amssymb, bm}

\subsection{Bi-Fidelity Gaussian Process Classification Model}

To model the binary classification problem with data from two fidelities, we follow the non-linear auto-regressive scheme originally proposed in \cite{kennedy2000predicting} and adapted for classification in \cite{costabal2019multi}.
The core of the model is built upon latent Gaussian Processes, but due to the Bernoulli likelihood used for binary data, exact inference is not possible. Unlike \cite{costabal2019multi}, we use a variational inference framework. 
This design choice was made for efficiency reasons, because the model inference step is repeated a large number of times in the active learning method we present.

\subsubsection{Model Specification}

Let $\mathcal{D}_L = \{(\mathbf{x}_{L,i}, y_{L,i})\}_{i=1}^{N_L}$ and $\mathcal{D}_H = \{(\mathbf{x}_{H,j}, y_{H,j})\}_{j=1}^{N_H}$ be the training datasets for the low- and high-fidelity sources, respectively, where $\mathbf{x} \in \mathbb{R}^d$ are input parameters and $y \in \{0, 1\}$ are the binary class labels. 
We introduce two latent functions, $f_L(\mathbf{x})$ and $f_H(\mathbf{x})$. 
The relationship between the observed binary labels and these latent functions is defined through a probit link function (the cumulative distribution function of the standard normal distribution, $\Phi(\cdot)$):
\begin{align}
    p(y_L=1 | \mathbf{x}) &= \Phi(f_L(\mathbf{x})) \\
    p(y_H=1 | \mathbf{x}) &= \Phi(f_H(\mathbf{x}))
\end{align}
The bi-fidelity structure is established through an auto-regressive model connecting the two latent functions:
\begin{equation}
    f_H(\mathbf{x}) = \rho f_L(\mathbf{x}) + \delta(\mathbf{x})
    \label{eq:autoregressive}
\end{equation}
where $\rho$ is a learnable scalar parameter that captures the correlation between the fidelities, and $\delta(\mathbf{x})$ is a discrepancy function that models the difference between the scaled low-fidelity function and the high-fidelity function.

\begin{figure}[t]
    \centering

\begin{tikzpicture}[node distance=2.2cm,>=Stealth,thick,]

% Nodes
\node[circle,draw,minimum size=1.2cm] (fL) {\(f_L\)};
\node[rectangle,draw,minimum size=1.2cm, right=2.7cm of fL] (yL) {\(y_L\)};
\node[circle,draw,minimum size=1.2cm, below=0.8cm of fL] (fH) {\(f_H\)};
\node[circle,draw,minimum size=1.2cm, left=2cm of fH] (delta) {\(\delta\)};
\node[rectangle,draw,minimum size=1.2cm, right=2.7cm of fH] (yH) {\(y_H\)};

% Arrows
\draw[->] (fL) -- (yL);
\draw[->] (delta) -- (fH);
\draw[->] (fH) -- (yH);
\draw[->] (fL) -- (fH);

% Node labels
\node[align=center] at ($(fH)+(0,-1)$) {\(f_H=\rho f_L + \delta\)};
\node[align=center] at ($(fL)+(0,1)$) {\(f_L \sim \mathcal{GP}\)};
\node[align=center] at ($(delta)+(0,-1)$) {\(\delta \sim \mathcal{GP}\)};
\node[align=center] at ($(yL)+(2,1)$) {\(P(y_L \mid f_L) = \Phi(f_L)^{y_L} (1 - \Phi(f_L))^{1-y_L}\)};
\node[align=center] at ($(yH)+(2,-1)$) {\(P(y_H \mid f_H) = \Phi(f_H)^{y_H} (1 - \Phi(f_H))^{1-y_H}\)};

% Vertical dashed dividing line
\draw [dashed] ($(fL)!0.5!(yL) + (0,1.25)$) -- ++(0.,-4.6);

% Horizontal braces and labels, BELOW the nodes
% Latents: from delta's left to fL's right, a bit below
%\draw [decorate,decoration={brace,mirror,amplitude=8pt}] 
%  ($(delta) + (-0.7,-1.5)$) -- ($(fH) + (0.7,-1.5)$)
%  node[midway, yshift=-12pt] {\large latents};

% Observables: from fH's left to yL's right, a bit below
%\draw [decorate,decoration={brace,mirror,amplitude=8pt}] 
%  ($(yH) + (-0.7,-1.5)$) -- ($(yH) + (0.7,-1.5)$)
%  node[midway, yshift=-12pt] {\large observables};

\end{tikzpicture}

    \caption{Bi-fidelity Gaussian Process classification model due to \cite{costabal2019multi} used in this work. Variables on the left of the dashed line are hidden (latents), while $y_L$ and $y_H$ are the binary observed outcomes.}
    \label{fig:bfgp}
\end{figure}
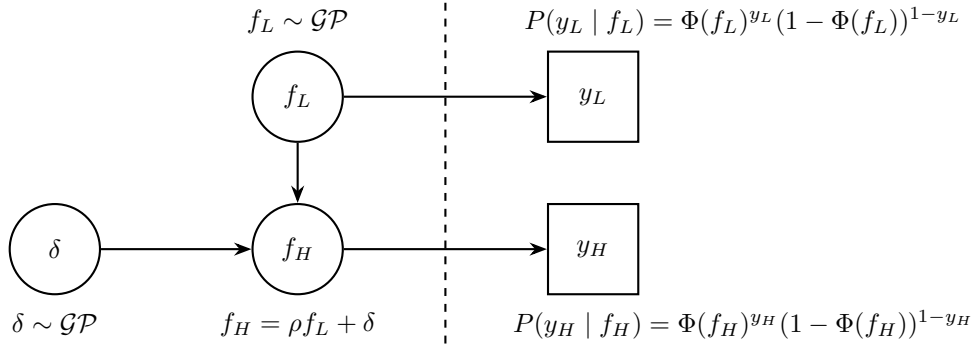

We place independent Gaussian Process priors on the low-fidelity latent function $f_L(\mathbf{x})$ and the discrepancy function $\delta(\mathbf{x})$:
\begin{align}
    f_L(\mathbf{x}) &\sim \mathcal{GP}(m_L(\mathbf{x}), k_L(\mathbf{x}, \mathbf{x}')) \\
    \delta(\mathbf{x}) &\sim \mathcal{GP}(m_\delta(\mathbf{x}), k_\delta(\mathbf{x}, \mathbf{x}'))
\end{align}
In our implementation, both GPs use a constant mean function, $m(\mathbf{x}) = c$, and a scaled Radial Basis Function (RBF) kernel:
\begin{equation}
    k(\mathbf{x}, \mathbf{x}') = \sigma^2 \exp\left(-\frac{\|\mathbf{x} - \mathbf{x}'\|^2}{2\ell^2}\right)
\end{equation}
where the output-scale $\sigma^2$ and the lengthscale $\ell$ are hyperparameters. Due to the independence of the priors on $f_L$ and $\delta$, the joint prior distribution over the latent functions is also a Gaussian Process.
Define the vector-valued function $\mathbf{f}(\mathbf{x}) = [f_L(\mathbf{x}), f_H(\mathbf{x})]^T$. This joint process is fully specified by its vector-valued mean function $\mathbf{m}(\mathbf{x})$ and its matrix-valued covariance function (or kernel matrix) $\mathbf{K}(\mathbf{x}, \mathbf{x}')$:
\begin{equation}
    \begin{bmatrix}
        f_L(\mathbf{x}) \\
        f_H(\mathbf{x})
    \end{bmatrix}
    \sim \mathcal{GP} \left( \mathbf{m}(\mathbf{x}), \mathbf{K}(\mathbf{x}, \mathbf{x}') \right),
\end{equation}
where the mean function is:
\begin{equation}
    \mathbf{m}(\mathbf{x}) =
    \begin{bmatrix}
        \mathbb{E}[f_L(\mathbf{x})] \\
        \mathbb{E}[f_H(\mathbf{x})]
    \end{bmatrix}
    =
    \begin{bmatrix}
        m_L(\mathbf{x}) \\
        \rho m_L(\mathbf{x}) + m_\delta(\mathbf{x})
    \end{bmatrix},
\end{equation}
and the covariance function is:
\begin{equation}
    \mathbf{K}(\mathbf{x}, \mathbf{x}') =
    \begin{bmatrix}
        k_L(\mathbf{x}, \mathbf{x}') & \rho k_L(\mathbf{x}, \mathbf{x}') \\
        \rho k_L(\mathbf{x}, \mathbf{x}') & \rho^2 k_L(\mathbf{x}, \mathbf{x}') + k_\delta(\mathbf{x}, \mathbf{x}')
    \end{bmatrix}.
\end{equation}
A graphical illustration of this model is provided in Figure \ref{fig:bfgp}.

\subsubsection{Variational Inference and Training}

We use a variational inference approach to find an approximate posterior, implemented in the \texttt{gpytorch} library \cite{gardner2018gpytorch}.
 We define variational distributions over the values of the latent functions at a set of inducing locations $\mathbf{Z}_L$ and $\mathbf{Z}_\delta$:
\begin{align}
    q(\mathbf{u}_L) &= \mathcal{N}(\mathbf{m}_{u_L}, \mathbf{S}_{u_L}) \quad \text{where} \quad \mathbf{u}_L = f_L(\mathbf{Z}_L) \\
    q(\mathbf{u}_\delta) &= \mathcal{N}(\mathbf{m}_{u_\delta}, \mathbf{S}_{u_\delta}) \quad \text{where} \quad \mathbf{u}_\delta = \delta(\mathbf{Z}_\delta)
\end{align}
Here, $\mathbf{m}$ and $\mathbf{S}$ are the variational parameters (mean vectors and Cholesky factors) to be optimized. 
Training consists of maximizing the Evidence Lower Bound (ELBO), $\mathcal{L}_{\text{ELBO}}$, with respect to the variational parameters, the model hyperparameters, and the scaling factor $\rho$. The ELBO is given by:
\begin{equation}
    \mathcal{L}_{\text{ELBO}} = \mathbb{E}_{q(f_L)}[\log p(\mathcal{D}_L | f_L)] + \mathbb{E}_{q(f_H)}[\log p(\mathcal{D}_H | f_H)] - \text{KL}[q(\mathbf{u}_L) || p(\mathbf{u}_L)] - \text{KL}[q(\mathbf{u}_\delta) || p(\mathbf{u}_\delta)]
\label{eq:elbo}
\end{equation}
where the expectation $\mathbb{E}_{q(f_H)}$ is taken over the distribution of $f_H$ induced by $q(f_L)$ and $q(\delta)$ via Equation \ref{eq:autoregressive}. The expected log-likelihood terms are computed using Gauss-Hermite quadrature. 

Additionally, to prevent overfitting an L2 regularization term is added to the objective function, which is equivalent to placing a broad Gaussian prior on the kernel hyperparameters and minimizing the negative log-posterior:
\begin{equation}
    \mathcal{L}_{\text{final}} = -\mathcal{L}_{\text{ELBO}} + \lambda \sum_{\theta \in \Theta} (\theta - \theta_{\text{prior}})^2
\end{equation}
where $\Theta$ is the set of all kernel hyperparameters and $\lambda$ is a regularization coefficient (default value $10^{-2}$). The model is trained by minimizing $\mathcal{L}_{\text{final}}$ using the Adam optimizer (learning rate $10^{-3}$). To improve robustness against local minima, the training process is repeated for three random initializations, and the model yielding the best final ELBO is selected.

\subsubsection{Prediction}

After training, the model can be used to make predictions at new test locations $\mathbf{x}^*$.
The approximate predictive posterior for the high-fidelity latent function, $q(f_H(\mathbf{x}^*))$, is constructed based on the auto-regressive model (Equation \ref{eq:autoregressive}):
\begin{align}
    \mathbb{E}[f_H(\mathbf{x}^*)] &= \rho \mathbb{E}[f_L(\mathbf{x}^*)] + \mathbb{E}[\delta(\mathbf{x}^*)] \\
    \text{Var}[f_H(\mathbf{x}^*)] &= \rho^2 \text{Var}[f_L(\mathbf{x}^*)] + \text{Var}[\delta(\mathbf{x}^*)]
\end{align}
The final predictive probability for the high-fidelity class is obtained by marginalizing over the uncertainty in this latent posterior:
\begin{equation}
    p(y_H=1 | \mathbf{x}^*, \mathcal{D}) = \int \Phi(f_H) q(f_H(\mathbf{x}^*)) df_H
\end{equation}
This integral is again approximated numerically.

\subsection{Batch acquisition function}

Active learning aims to intelligently select new data points for labeling to improve model performance under a limited budget. 
In the \textit{batched, multi-fidelity} setting, the goal is to select multiple queries without re-training the model, where each query is an (input location, fidelity) pair, $(\mathbf{x}, m)$, with $m \in \{L, H\}$. 
Let $c_L$ and $c_H$ be the costs associated with acquiring a low- or high-fidelity label, respectively. 
Given a total budget $B$ for an active learning step, we seek to find the batch of queries $\mathcal{Q} = \{(\mathbf{x}_i, m_i)\}_{i=1}^k$ that maximizes our objective, subject to the constraint that $\sum_{i=1}^k c_{m_i} \le B$.

\subsubsection{Mutual Information as an Acquisition Function}

We adopt an information-theoretic acquisition function based on Mutual Information (MI), following the batched multi-fidelity active learning approach in \cite{li2022batch}. A variety of different MI-based acquisition functions are developed and compared in this work, and these are defined later. For now, denote the acquisition function as $\alpha(\mathcal{Q})$. Finding the optimal batch is a combinatorial NP-hard problem. However, mutual information is known to be a \textit{submodular} function. 

A set function $F$ is submodular if it exhibits a diminishing returns property: for any sets $A \subseteq B$ and an element $x \notin B$, the marginal gain of adding $x$ to $B$ is no more than adding it to $A$, i.e., $F(A \cup \{x\}) - F(A) \ge F(B \cup \{x\}) - F(B)$. 
For maximizing a monotonic submodular function under a budget constraint, a simple greedy algorithm is guaranteed to find a solution that is within a constant factor $(1 - 1/e)$ of the optimal solution \cite{nemhauser1978analysis}.

This allows us to construct the batch sequentially. Starting with an empty batch $\mathcal{Q}_0 = \emptyset$, at each step $k$, we add the single query $(\mathbf{x}^*, m^*)$ that offers the largest marginal information gain per unit cost:
\begin{equation}
    (\mathbf{x}^*, m^*) = \arg\max_{(\mathbf{x}, m)} \frac{\alpha(\mathcal{Q}_k \cup \{(\mathbf{x}, m)\}) - \alpha(\mathcal{Q}_k)}{c_m}
\end{equation}
This process is repeated until the budget is exhausted. As discussed in \cite{li2022batch}, in the multi-fidelity setting there is a minor caveat. The algorithm continues until a proposal is found which will exceed the budget; we select this and then terminate. This preserves the near-optimality property in the multi-fidelity setting.

\subsubsection{Baseline: Latent Function Mutual Information (LFMI)}
\label{sec:lfmi}

Directly adapting the approach in \cite{li2022batch} to our bi-fidelity Gaussian Process model yields an acquisition function designed to maximize the information gained about the high-fidelity latent function $f_H$ across the entire input domain. This is practically estimated by computing the mutual information between the latent function values at the candidate batch $\mathcal{Q}$ and the high-fidelity latent function values at a fixed set of test locations $\mathcal{X}' = \{\mathbf{x}'_j\}_{j=1}^{N'}$, which are sampled uniformly from the domain to represent the space over which we want to reduce uncertainty. The acquisition function is thus the joint mutual information:
\begin{equation}
    \alpha_{\text{LFMI}}(\mathcal{Q}) = I(\mathbf{f}_{\mathcal{Q}}; \mathbf{f}_{H, \mathcal{X}'}),
    \label{eq:lfmi_acquisition}
\end{equation}
where $\mathbf{f}_{\mathcal{Q}}$ is the vector of latent function values corresponding to the queries in $\mathcal{Q}$ (i.e., $f_L(\mathbf{x}_i)$ if $m_i=L$ and $f_H(\mathbf{x}_i)$ if $m_i=H$), and $\mathbf{f}_{H, \mathcal{X}'}$ are the latent values of the high-fidelity function at the test locations $\mathcal{X}'$.

For a joint Gaussian distribution, this MI has a convenient analytical form based on the log-determinants of the covariance matrices:
\begin{equation}
    I(\mathbf{A}; \mathbf{B}) = \frac{1}{2} \left( \log\det(\bm{\Sigma}_{\mathbf{A}}) + \log\det(\bm{\Sigma}_{\mathbf{B}}) - \log\det(\bm{\Sigma}_{\mathbf{A},\mathbf{B}}) \right)
\end{equation}
where $\bm{\Sigma}_{\mathbf{A}}$, $\bm{\Sigma}_{\mathbf{B}}$, and $\bm{\Sigma}_{\mathbf{A},\mathbf{B}}$ are the covariance matrices of the random vectors $\mathbf{A}$, $\mathbf{B}$, and their joint distribution, respectively. 

These quantities can be efficiently computed from our bi-fidelity GPC model as follows. We construct the joint posterior distribution over the target vector $\mathbf{f}_{\text{joint}} = [\mathbf{f}_{\mathcal{Q}}^T, \mathbf{f}_{H, \mathcal{X}'}^T]^T$ from the base independent posteriors $q(f_L)$ and $q(\delta)$. Let $\mathcal{X}_{L}^{\text{eval}}$ be the set of unique input locations required for all $f_L$ evaluations, and $\mathcal{X}_{\delta}^{\text{eval}}$ be the set for all $\delta$ evaluations. 
We form a base random vector $\mathbf{g}$ by concatenating the latent posteriors at these unique points:
\begin{equation}
    \mathbf{g} = 
    \begin{bmatrix}
        \mathbf{f}_L(\mathcal{X}_{L}^{\text{eval}}) \\
        \bm{\delta}(\mathcal{X}_{\delta}^{\text{eval}})
    \end{bmatrix}
\end{equation}
Due to the independence of the variational posteriors for $f_L$ and $\delta$, the distribution of this base vector is a block-diagonal Gaussian:
\begin{equation}
    \mathbf{g} \sim \mathcal{N}\left( 
    \begin{bmatrix}
        \bm{\mu}_{f_L} \\ \bm{\mu}_{\delta}
    \end{bmatrix}, 
    \begin{bmatrix}
        \bm{\Sigma}_{f_L} & \mathbf{0} \\
        \mathbf{0} & \bm{\Sigma}_{\delta}
    \end{bmatrix}
    \right),
\end{equation}
where $\bm{\mu}_{f_L}$ and $\bm{\Sigma}_{f_L}$ are the mean and covariance of $q(f_L(\mathcal{X}_L^{\text{eval}}))$, and likewise for $\delta$.

The target vector $\mathbf{f}_{\text{joint}}$ is a linear transformation of this base vector, $\mathbf{f}_{\text{joint}} = \mathbf{T} \mathbf{g}$. The transformation matrix $\mathbf{T}$ assembles the final outputs from the base components according to Equation~\ref{eq:autoregressive}. It takes the block matrix form:
\begin{equation}
    \mathbf{T} = 
    \begin{bmatrix}
        \mathbf{S}_{L,Q} & \mathbf{0} \\
        \rho \mathbf{S}_{L,H} & \mathbf{S}_{\delta,H} \\
        \rho \mathbf{S}_{L,X'} & \mathbf{S}_{\delta,X'}
    \end{bmatrix}
\end{equation}
Here, the $\mathbf{S}$ matrices are sparse selection matrices of 0s and 1s. For instance, $\mathbf{S}_{L,Q}$ maps the entries of $\mathbf{f}_L(\mathcal{X}_L^{\text{eval}})$ to their corresponding positions in the low-fidelity query portion of $\mathbf{f}_{\text{joint}}$. The first block row constructs the low-fidelity query outputs ($f_L$), while the second and third rows construct the high-fidelity outputs ($f_H = \rho f_L + \delta$) for the high-fidelity queries and test points, respectively. The resulting distribution of the target vector is also Gaussian, $q(\mathbf{f}_{\text{joint}}) = \mathcal{N}(\mathbf{T}\bm{\mu}_g, \mathbf{T}\bm{\Sigma}_g\mathbf{T}^T)$, from which all necessary covariance matrices for the MI calculation can be extracted.

However, this approach has a significant limitation for classification and probability estimation tasks. The probit function used to transform the unbounded latents into a probability saturates if $f_H \gg 0$ or $f_H \ll 0$. This is not accounted for when measuring MI between the latent functions. The LFMI acquisition function may favor acquiring points in regions where the model is already highly confident, as the latent uncertainty can still be large.
This is inefficient, as reducing uncertainty in these saturated regions does not improve knowledge of the Bernoulli parameter.

\subsubsection{Proposed Method: Bernoulli Parameter Mutual Information (BPMI)}
\label{sec:proposed_mi}

To address the saturation issue, we propose a novel acquisition function that directly approximates the information gain with respect to the classification probabilities, which we denote as Bernoulli parameters $\mathbf{p} = \Phi(\mathbf{f})$. 
The objective is to maximize $I(\mathbf{p}_{\mathcal{Q}}; \mathbf{p}_{H, \mathcal{X}'})$.
Since the probit transformation $\Phi(\cdot)$ is non-linear, the distribution of $\mathbf{p}$ is non-Gaussian, and its MI is intractable.
We circumvent this issue by applying a first-order Taylor expansion of the link function around the posterior mean of the latent functions, thereby linearizing the link function and enabling an efficient approximation of $I(\mathbf{p}_{\mathcal{Q}}; \mathbf{p}_{H, \mathcal{X}'})$. We now describe this procedure in more detail.

Let $\mathbf{f}_{\text{joint}} = [\mathbf{f}_{\mathcal{Q}}, \mathbf{f}_{H, \mathcal{X}'}]^T$ be the joint latent vector with posterior $q(\mathbf{f}_{\text{joint}}) = \mathcal{N}(\bm{\mu}_f, \bm{\Sigma}_f)$. The corresponding vector of probabilities $\mathbf{p}_{\text{joint}}$ is approximated as:
\begin{equation}
    \mathbf{p}_{\text{joint}} \approx \Phi(\bm{\mu}_f) + (\mathbf{f}_{\text{joint}} - \bm{\mu}_f) \odot \Phi'(\bm{\mu}_f)
\end{equation}
where $\odot$ is the element-wise product and $\Phi'(\cdot)$ is the derivative of the probit link function, which is the standard normal probability density function, $\phi(\cdot)$.
This linearization leads to a Gaussian approximation for the distribution of the Bernoulli parameters, $q(\mathbf{p}_{\text{joint}}) \approx \mathcal{N}(\bm{\mu}_p, \bm{\Sigma}_p)$, where:
\begin{align}
    \bm{\mu}_p &= \Phi(\bm{\mu}_f) \\
    \bm{\Sigma}_p &\approx \mathbf{D} \bm{\Sigma}_f \mathbf{D}^T
    \label{eq:linearized_cov}
\end{align}
and $\mathbf{D}$ is a diagonal matrix containing the derivatives evaluated at the mean: $\mathbf{D} = \text{diag}(\phi(\bm{\mu}_f))$.

The MI can now be calculated analytically using this approximated Gaussian distribution over the probabilities. The key advantage of this formulation is that the derivative term $\phi(\mu_f)$ approaches zero as $|\mu_f|$ becomes large. Consequently, the elements of the transformed covariance matrix $\bm{\Sigma}_p$ will be small in regions where the classification probability is already saturated (close to 0 or 1). This naturally focuses the acquisition function on uncertain regions near the decision boundary, where the potential information gain is highest, leading to more efficient learning.

\subsubsection{Sampling frequency}
\label{sec:repeats}

For problems with a Bernoulli-distributed outcome, a single observation at a location $\mathbf{x}$ provides limited information about the underlying success probability $p(\mathbf{x})$. To gain a more precise estimate of $p(\mathbf{x})$, it is advantageous to perform multiple trials at or near the same location. 
Our proposed BPMI method also adaptively determines the number of times a selected point should be sampled, which we refer to as the sampling frequency or number of repeats, $N$.

We propose a heuristic that bases the number of repeats on the model's current estimate of the aleatoric (data) uncertainty. 
The standard error of the sample mean from $N$ samples of a Bernoulli random variable is $\sqrt{\frac{p(1-p)}{N}}$. To achieve a fixed standard error, $N$ must be proportional to $p(1-p)$, which is maximised at $p=0.5$.
This further concentrates sampling in regions of maximum aleatoric uncertainty ($p \approx 0.5$).

When a point $\mathbf{x}$ is chosen by the greedy acquisition strategy, we first use the current model to predict the mean probability, $p_{\text{pred}} = \mathbb{E}[p(\mathbf{x})]$. The number of repeats $N$ is then calculated as:
\begin{equation}
    N(p_{\text{pred}}) = \text{round}\left( \frac{N_{max}-1}{4} \cdot (p_{\text{pred}}(1-p_{\text{pred}}) + 1)  \right)
\end{equation}
where $N_{\text{max}}$ is a hyperparameter defining the maximum number of repeats allowed. 
Finally, as a practical implementation detail, a small amount of random jitter is added to the input coordinates of the repeated samples to avoid placing multiple query points at the exact same location, which can cause numerical issues (non-positive definiteness) in the GP model.

\subsubsection{Additional baselines}

We also compare our method against two non-MI baselines.

\begin{enumerate}
    \item \textbf{Random Point Selection} is a simple strategy that does not use any information from the model. To form a batch, it sequentially selects a random point from each fidelity to query until the step's budget is spent. 

    \item \textbf{Maximum Uncertainty Point Selection} is a common active learning heuristic that queries points where the model is least confident. The acquisition score is a weighted sum of two uncertainty types. The first is epistemic uncertainty (model uncertainty), captured by the variance of the model’s prediction for the Bernoulli success probability p. The second is aleatoric uncertainty (data uncertainty), captured by the entropy of the predicted Bernoulli distribution, which is highest when the predicted probability is 0.5. The relative importance of these two terms is controlled by a hyperparameter $\beta$ (default 0.5). The strategy greedily selects one point from each fidelit until the budget is filled.
    
\end{enumerate}

\section{Numerical Experiments}

\subsection{Toy problems}

\begin{figure}
    \centering
    \begin{subfigure}[b]{0.49\textwidth}
        \includegraphics[width=\textwidth]{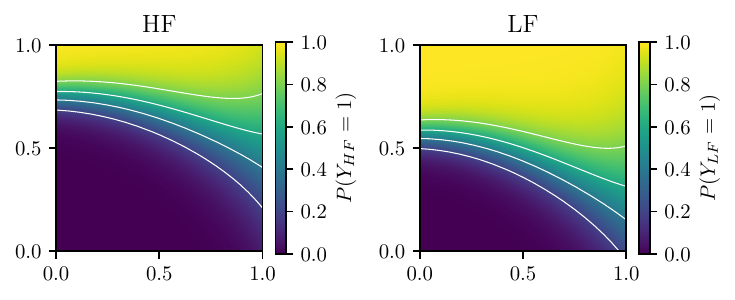}
        \caption{Linear discrepancy beween fidelities}
    \end{subfigure}
    \begin{subfigure}[b]{0.49\textwidth}
        \includegraphics[width=\textwidth]{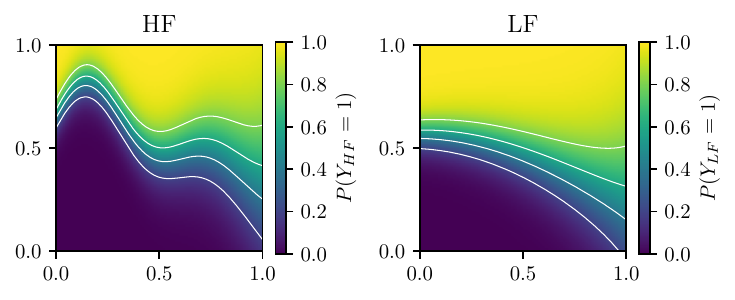}
        \caption{Nonlinear discrepancy between fidelities}
    \end{subfigure}
    \caption{Toy problems used in numerical experiments, showing the probability as a function of a 2D parameter space. Observed data at a given point is the outcome of Bernoulli trial using the probability at that point illustrated above.}
    \label{fig:toy}
\end{figure}

We introduce two synthetic 2D binary classification problems to simulate multi-fidelity data. The input space for both problems is the unit square, $\mathbf{x} = (x_1, x_2) \in [0, 1]^2$. 
We define a probability field $p(\boldsymbol x)$, and data at $x$ is sampled according to $y(\boldsymbol x) \sim \mathrm{Bernoulli}(p(\boldsymbol x))$. These probability fields are illustrated in Figure \ref{fig:toy}.

The low-fidelity data source, $y_L$, is common to both problems. It is defined by:
\begin{align}
    P(y_L=1|\mathbf{x}) &= \sigma\left(s(x_1) \cdot (x_2 - d_L(x_1))\right) \\
    \text{where } d_L(x_1) &= \frac{1}{3} \left( \cos\left(\frac{\pi x_1}{2}\right) + 1 \right) - 0.1 \label{eq:lf_boundary}\\
    \text{and } s(x_1) &= \alpha (1 - 0.75 x_1) \label{eq:scale}
\end{align}
Here, $\sigma(z) = (1+e^{-z})^{-1}$ is the sigmoid function and $\alpha$ is a base scale factor (set to $\alpha=20$) that controls the overall sharpness of the boundary.

The high-fidelity source, $y_H$, is defined by transforming the LF decision boundary. We consider two scenarios:

\paragraph{1. Linear Transformation}
The HF decision boundary $d_{H, \text{lin}}$ is a linear shift and scale of the LF boundary. The resulting probability is:
\begin{align}
    P(y_{H, \text{lin}}=1|\mathbf{x}) &= \sigma\left(s(x_1) \cdot (x_2 - d_{H, \text{lin}}(x_1))\right) \\
    \text{where } d_{H, \text{lin}}(x_1) &= 0.8 \, d_L(x_1) + 0.3 \label{eq:hf_linear}
\end{align}

\paragraph{2. Nonlinear Transformation}
The HF decision boundary $d_{H, \text{nlin}}$ is a more complex, nonlinear transformation of the LF boundary. The probability is:
\begin{align}
    P(y_{H, \text{nlin}}=1|\mathbf{x}) &= \sigma\left(s(x_1) \cdot (x_2 - d_{H, \text{nlin}}(x_1))\right) \\
    \text{where } d_{H, \text{nlin}}(x_1) &= d_L(x_1) + 0.2 \sin(3\pi x_1)(1 - x_1) + 0.1 \label{eq:hf_nonlinear}
\end{align}

In our toy problems, we make the following design choices, which mimic our practical application. The cost ratio of the fidelities is set to 10 ($c_{LF}=0.1$, $c_{HF}=1$). We initialize our model using 50 LF and 25 HF runs, and then employ 5 active learning rounds, with a batch cost of 100 per round. Each algorithm is repeated 20 times from different initial data, which was sufficient to obtain converged statistics. The random seed is set such that for a given repeat each algorithm starts from identical data.

\paragraph{Metrics}
We assess the performance of each active learning strategy by evaluating the model on a fixed test set of 10,000 points drawn from the high-fidelity distribution. We use two primary metrics.

First, we compute the Expected Log Predictive Probability (ELPP), a standard metric for evaluating probabilistic models. It measures the average log-likelihood of the high-fidelity test data $\mathcal{D}_{\text{test}} = \{(\mathbf{x}_i, y_{H,i})\}_{i=1}^{N_{\text{test}}}$ under the model's predictive distribution. For a binary classification model that predicts the probability $P(Y_H=1|\mathbf{x}_i)$ as $\hat{p}_H(\mathbf{x}_i)$, the ELPP is defined as:
\begin{equation}
    \text{ELPP} = \frac{1}{N_{\text{test}}} \sum_{i=1}^{N_{\text{test}}} \left[ y_{H,i} \log \hat{p}_H(\mathbf{x}_i) + (1-y_{H,i}) \log(1-\hat{p}_H(\mathbf{x}_i)) \right].
    \label{eq:elpp}
\end{equation}
Higher values indicated better performance.

Second, because we are in a synthetic setting where the true underlying probability field $p_{H, \text{true}}(\mathbf{x})$ is known, we can directly measure the model's accuracy in reconstructing this field. We compute the Mean Squared Error (MSE) between the model's predicted probability $\hat{p}_H(\mathbf{x})$ and the true probability $p_{H, \text{true}}(\mathbf{x})$ over the test set:
\begin{equation}
    \text{MSE} = \frac{1}{N_{\text{test}}} \sum_{i=1}^{N_{\text{test}}} \left( \hat{p}_H(\mathbf{x}_i) - p_{H, \text{true}}(\mathbf{x}_i) \right)^2.
    \label{eq:mse}
\end{equation} 
Lower values indicate better performance.

\begin{figure}
    \centering
    \begin{subfigure}[b]{0.49\textwidth}
        \includegraphics[width=\textwidth]{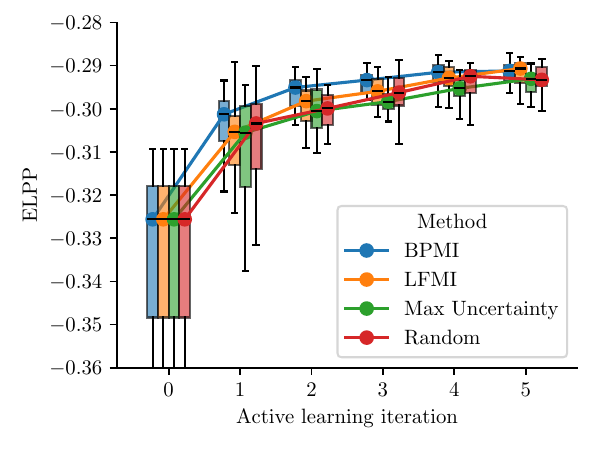}
        \caption{Linear toy problem ELPP}
    \end{subfigure}
    \begin{subfigure}[b]{0.49\textwidth}
        \includegraphics[width=\textwidth]{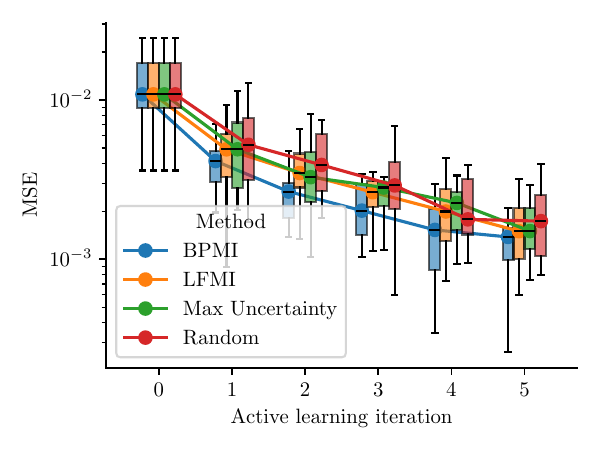}
        \caption{Linear toy problem MSE}
    \end{subfigure}

        \begin{subfigure}[b]{0.49\textwidth}
        \includegraphics[width=\textwidth]{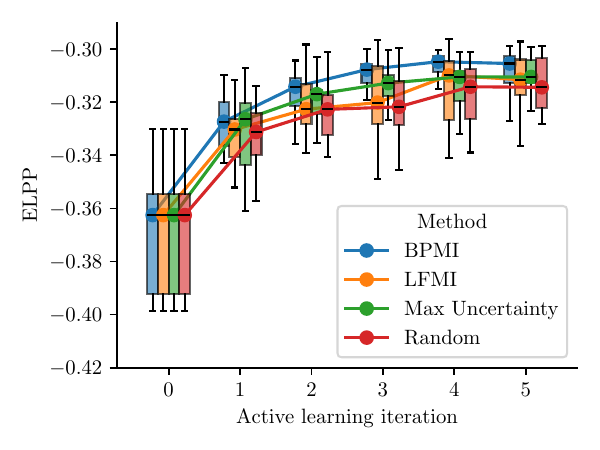}
        \caption{Nonlinear toy problem ELPP}
    \end{subfigure}
    \begin{subfigure}[b]{0.49\textwidth}
        \includegraphics[width=\textwidth]{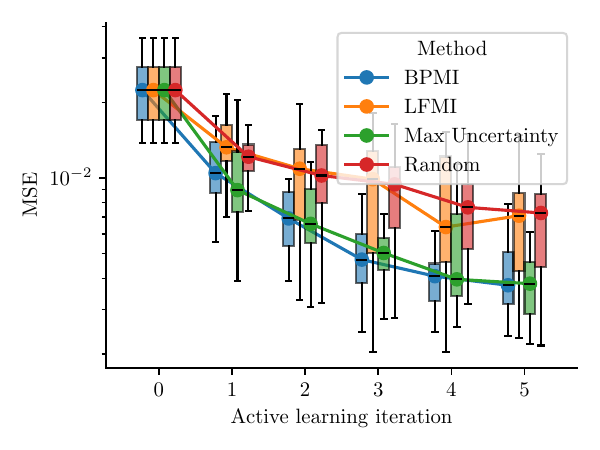}
        \caption{Nonlinear toy problem MSE}
    \end{subfigure}
    
    \caption{Comparison of active learning algorithms on the toy problems. BPMI consistently achieves higher ELPP and lower MSE than baseline methods across 20 independent runs.}
    \label{fig:toyresults}
\end{figure}

\paragraph{Results} Batch active learning algorithms are compared on these two toy problems in Figure \ref{fig:toyresults}. The proposed BPMI algorithm is consistently superior. However, in future work we plan to expand our evaluation problem set, which at present is limited.

\subsection{Laser-ignited rocket combustor application}

The motivating application of this work is a laser-ignited rocket combustor.
Laser ignition is an ignition approach which could facilitate unlimited engine re-ignition. We consider a circular jet of oxygen with an annular flow of methane, and study the effect on ignition when the laser focal location is varied throughout the combustion chamber. In certain regions the laser spark interacts with the turbulent shear layer of the jet, and in these cases the ignition outcome can depend on the instantaneous turbulence, meaning that the ignition outcome is random (with an unknown underlying probability). We wish to quantify the ignition probability over a two-dimensional plane of symmetry through the combustor centreline.

% Add schematic of the comustor / MF setup (two schlierens?)

\paragraph{Multi-fidelity methodology}

The multi-fidelity methodology is described in \cite{cutforth2025bi}, and briefly summarized here for completeness. 
Simulations are carried out using the Hypersonics Task-based Research (HTR) solver \cite{di2020htr}, modeling the compressible chemically reacting Navier-Stokes equations. The computational details concerning the solver parallelism, simulation domain, boundary conditions, and numerics can be found in previous work as part of PSAAP-III \cite{wang2021progress, passiatore2024computational, zahtila2025bi}. Note that the laser is modeled by an energy source term imposed on the fluid.

Two mesh resolutions are employed in this study. The low-fidelity mesh is composed of 2M grid points, and the high-fidelity counterpart is composed of 15M grid points. 
The same sub-grid scale models are employed in both cases \cite{smagorinsky1963general}. A fully resolved direct numerical simulation would require approximately 100 billion grid points by comparison, but %presently the PSAAP project has achieved 
satisfactory agreement has been achieved between measured quantities in experiments \cite{strelau2023modes} and output statistics from the present resolution used in high-fidelity simulations. The chemistry in the low-fidelity simulations is simplified to a 5-species, 3-step mechanism. 
Simulations were carried out on a cluster consisting of 4 Nvidia V100 GPUs per node. Each simulation on the 15M mesh took approximately 128 GPU-hours, while the 2M mesh used approximately 8 GPU-hours, a cost ratio of approximately 16.

\paragraph{Evaluation method} 

We ran 491 HF simulations, focused on the transition region where the ignition probability switches from 0 to 1. We use these observed outcomes as a test set and evaluate the ELPP given in Equation \ref{eq:elpp} using these points.
Given computational constraints, we only compare our proposed BPMI active learning algorithm with our random selection baseline. We ran 3 active learning iterations, with a budget of 100 high-fidelity simulations per iteration (note that this budget is split between low- and hogh-fidelity simulations, so the actual number of high-fidelity runs per iteration is less than 100). Each active learning iteration therefore represents 12,800 GPU hours of compute.

% TODO: in future, get uncertainty on this result by bootstrap resampling the test set?

\begin{figure}
    \centering
    \begin{subfigure}[b]{0.69\textwidth}
        \includegraphics[width=\textwidth]{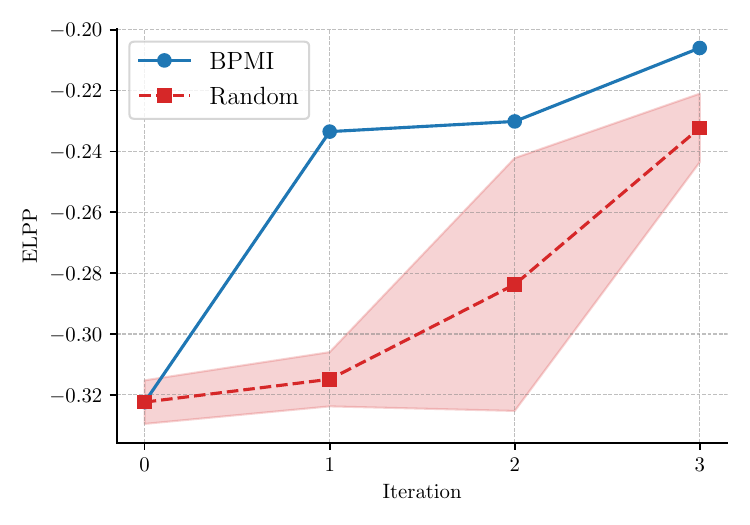}
        \caption{Test set ELPP versus batch active learning iteration}
    \end{subfigure}
    
    \begin{subfigure}[b]{0.24\textwidth}
        \includegraphics[width=\textwidth]{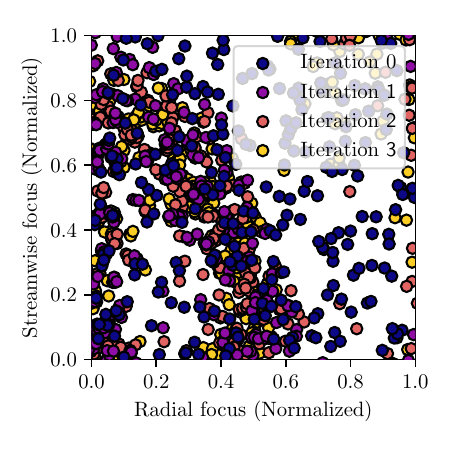}
        \caption{BPMI low-fidelity}
    \end{subfigure}
        \begin{subfigure}[b]{0.24\textwidth}
        \includegraphics[width=\textwidth]{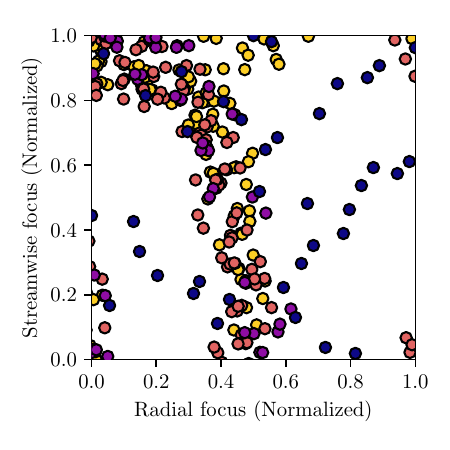}
        \caption{BPMI high-fidelity}
    \end{subfigure}
        \begin{subfigure}[b]{0.24\textwidth}
        \includegraphics[width=\textwidth]{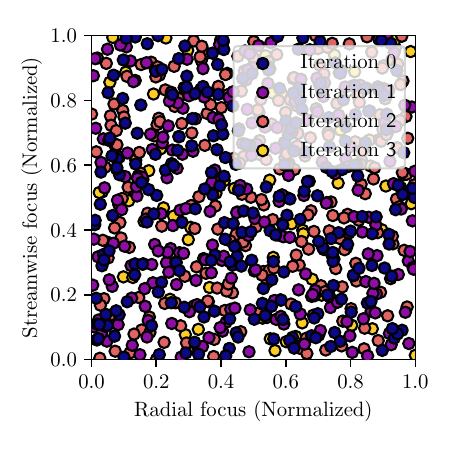}
        \caption{Random low-fidelity}
    \end{subfigure}
        \begin{subfigure}[b]{0.24\textwidth}
        \includegraphics[width=\textwidth]{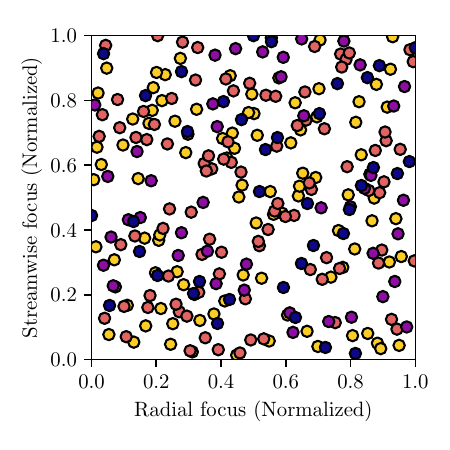}
        \caption{Random high-fidelity}
    \end{subfigure}

        \begin{subfigure}[b]{0.24\textwidth}
        \includegraphics[width=\textwidth]{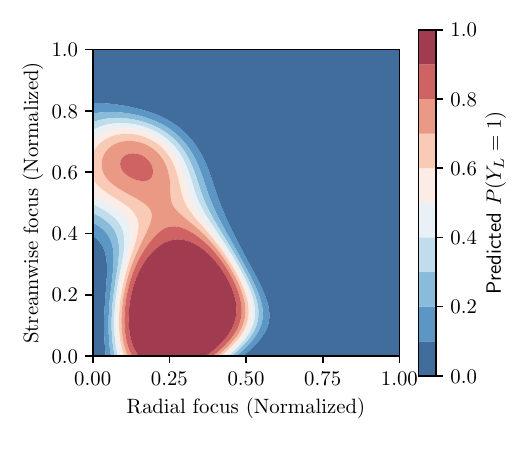}
        \caption{BPMI low-fidelity}
    \end{subfigure}
        \begin{subfigure}[b]{0.24\textwidth}
        \includegraphics[width=\textwidth]{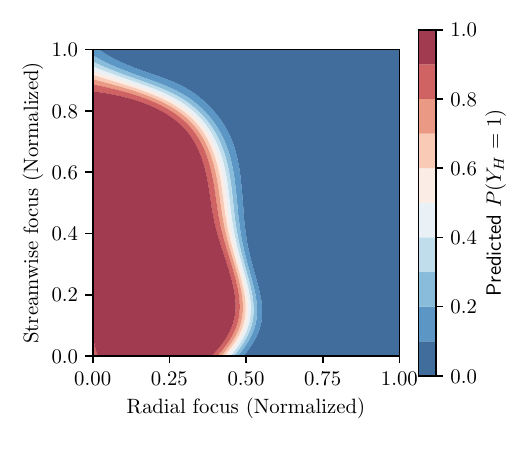}
        \caption{BPMI high-fidelity}
    \end{subfigure}
        \begin{subfigure}[b]{0.24\textwidth}
        \includegraphics[width=\textwidth]{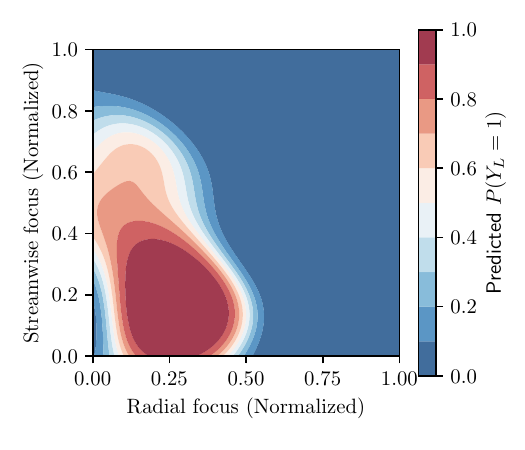}
        \caption{Random low-fidelity}
    \end{subfigure}
        \begin{subfigure}[b]{0.24\textwidth}
        \includegraphics[width=\textwidth]{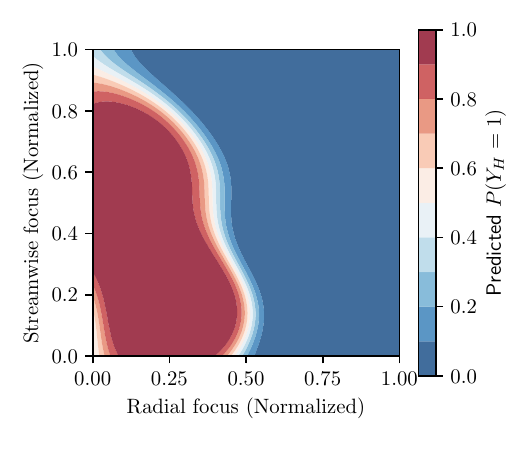}
        \caption{Random high-fidelity}
    \end{subfigure}

    \caption{Application of proposed active learning method to a complex multi-physics solver. In (a) the ELPP over a held-out test set is shown. The random strategy is repeated three times, with the shaded region representing the standard deviation of these repeats. Due to computational limitations the BPMI strategy was only run once. Panels (b) to (e) show the points selected for simulations by each approach. Panels (f) to (i) show the predicted ignition probabilities of the final model for each fidelity under the two active learning algorithms compared here.}
    \label{fig:psaapresults}
\end{figure}

\paragraph{Results} The results from applying these active learning problems to the laser-ignited combustor are shown in Figure \ref{fig:psaapresults}. Once again, for a given number of active learning iterations, the proposed BPMI method results in a better performing model. The reason for this superior performance is evident in the sampling patterns shown in panels (b-e). BPMI concentrates both low- and high-fidelity queries near the transition boundary (while automatically balancing exploration of the rest of the parameter domain), whereas the Random strategy disperses them.

\section{Conclusions}

In this work, we introduced a novel batch active learning strategy, Bernoulli Parameter Mutual Information (BPMI),
for binary data with a multi-fidelity Gaussian Process classification model. Our proposed method uses a linearization of the link function in used in Gaussian Process classification models in order to efficiently approximate a mutual information-based acquisition function.
This formulation naturally focuses sampling on the decision boundary, where uncertainty in the predicted probability is highest and information gain is most valuable. We further propose a heuristic for adaptively determining the number of repeated samples at a query location, based on the current estimate of aleatoric uncertainty.

Through numerical experiments on both synthetic benchmarks and a complex, real-world application of a laser-ignited rocket combustor, we demonstrated the superior performance of BPMI. Compared to an equivalent latent-space MI approach, a maximum uncertainty heuristic, and random sampling, our method consistently produced more accurate predictive models for a given computational budget. The results confirm that BPMI more effectively allocates resources to delineate the classification boundary, leading to faster convergence and a better final model. This work provides an effective and efficient framework for the sequential design of experiments in bi-fidelity, binary-outcome problems.

% \newpage
\section*{Acknowledgments}

The authors acknowledge financial support from the US Department of Energy’s National Nuclear Security Administration via the Stanford PSAAP-III Center for the prediction of laser ignition of a rocket combustor (DE-NA0003968).

%The authors also extend their gratitude to Dr.\ Gianluca Geracci, Prof.\ Gianluca Iaccarino, Dr.\ Marta D'Elia, and Dr.\ Juan Cardenas for helpful discussion over the course of this work.

%------------------------------------------------------------------------------

\bibliographystyle{unsrtnat}
\bibliography{references}

\begin{thebibliography}{18}
\providecommand{\natexlab}[1]{#1}
\providecommand{\url}[1]{\texttt{#1}}
\expandafter\ifx\csname urlstyle\endcsname\relax
  \providecommand{\doi}[1]{doi: #1}\else
  \providecommand{\doi}{doi: \begingroup \urlstyle{rm}\Url}\fi

\bibitem[Nickisch et~al.(2008)Nickisch, Rasmussen,
  et~al.]{nickisch2008approximations}
Hannes Nickisch, Carl~Edward Rasmussen, et~al.
\newblock Approximations for binary gaussian process classification.
\newblock \emph{Journal of Machine Learning Research}, 9\penalty0
  (10):\penalty0 2035--2078, 2008.

\bibitem[Kennedy and O'Hagan(2000)]{kennedy2000predicting}
Marc~C Kennedy and Anthony O'Hagan.
\newblock Predicting the output from a complex computer code when fast
  approximations are available.
\newblock \emph{Biometrika}, 87\penalty0 (1):\penalty0 1--13, 2000.

\bibitem[Costabal et~al.(2019)Costabal, Perdikaris, Kuhl, and
  Hurtado]{costabal2019multi}
Francisco~Sahli Costabal, Paris Perdikaris, Ellen Kuhl, and Daniel~E Hurtado.
\newblock Multi-fidelity classification using gaussian processes: accelerating
  the prediction of large-scale computational models.
\newblock \emph{Computer Methods in Applied Mechanics and Engineering},
  357:\penalty0 112602, 2019.

\bibitem[Jones et~al.(1998)Jones, Schonlau, and Welch]{jones1998efficient}
Donald~R Jones, Matthias Schonlau, and William~J Welch.
\newblock Efficient global optimization of expensive black-box functions.
\newblock \emph{Journal of Global optimization}, 13\penalty0 (4):\penalty0
  455--492, 1998.

\bibitem[Krause et~al.(2008)Krause, Singh, and Guestrin]{krause2008near}
Andreas Krause, Ajit Singh, and Carlos Guestrin.
\newblock Near-optimal sensor placements in gaussian processes: Theory,
  efficient algorithms and empirical studies.
\newblock \emph{Journal of Machine Learning Research}, 9\penalty0 (2), 2008.

\bibitem[Beck and Guillas(2016)]{beck2016sequential}
Joakim Beck and Serge Guillas.
\newblock Sequential design with mutual information for computer experiments
  (mice): Emulation of a tsunami model.
\newblock \emph{SIAM/ASA Journal on Uncertainty Quantification}, 4\penalty0
  (1):\penalty0 739--766, 2016.

\bibitem[Takeno et~al.(2020)Takeno, Fukuoka, Tsukada, Koyama, Shiga, Takeuchi,
  and Karasuyama]{takeno2020multi}
Shion Takeno, Hitoshi Fukuoka, Yuhki Tsukada, Toshiyuki Koyama, Motoki Shiga,
  Ichiro Takeuchi, and Masayuki Karasuyama.
\newblock Multi-fidelity bayesian optimization with max-value entropy search
  and its parallelization.
\newblock In \emph{International Conference on Machine Learning}, pages
  9334--9345. PMLR, 2020.

\bibitem[Kirsch et~al.(2019)Kirsch, Van~Amersfoort, and
  Gal]{kirsch2019batchbald}
Andreas Kirsch, Joost Van~Amersfoort, and Yarin Gal.
\newblock Batchbald: Efficient and diverse batch acquisition for deep bayesian
  active learning.
\newblock \emph{Advances in neural information processing systems}, 32, 2019.

\bibitem[Li et~al.(2022)Li, Phillips, Yu, Kirby, and Zhe]{li2022batch}
Shibo Li, Jeff~M Phillips, Xin Yu, Robert Kirby, and Shandian Zhe.
\newblock Batch multi-fidelity active learning with budget constraints.
\newblock \emph{Advances in Neural Information Processing Systems},
  35:\penalty0 995--1007, 2022.

\bibitem[Gardner et~al.(2018)Gardner, Pleiss, Bindel, Weinberger, and
  Wilson]{gardner2018gpytorch}
Jacob~R Gardner, Geoff Pleiss, David Bindel, Kilian~Q Weinberger, and
  Andrew~Gordon Wilson.
\newblock Gpytorch: Blackbox matrix-matrix gaussian process inference with gpu
  acceleration.
\newblock In \emph{Advances in Neural Information Processing Systems}, 2018.

\bibitem[Nemhauser et~al.(1978)Nemhauser, Wolsey, and
  Fisher]{nemhauser1978analysis}
George~L Nemhauser, Laurence~A Wolsey, and Marshall~L Fisher.
\newblock An analysis of approximations for maximizing submodular set
  functions—i.
\newblock \emph{Mathematical programming}, 14\penalty0 (1):\penalty0 265--294,
  1978.

\bibitem[Cutforth et~al.(2025)Cutforth, Fan, Zahtila, Doostan, and
  Darve]{cutforth2025bi}
Murray Cutforth, Tiffany Fan, Tony Zahtila, Alireza Doostan, and Eric Darve.
\newblock Bi-fidelity interpolative decomposition for multimodal data.
\newblock \emph{Submitted to Computer Methods in Applied Mechanics and
  Engineering}, 2025.

\bibitem[Di~Renzo et~al.(2020)Di~Renzo, Fu, and Urzay]{di2020htr}
Mario Di~Renzo, Lin Fu, and Javier Urzay.
\newblock Htr solver: An open-source exascale-oriented task-based multi-gpu
  high-order code for hypersonic aerothermodynamics.
\newblock \emph{Computer Physics Communications}, 255:\penalty0 107262, 2020.

\bibitem[Wang et~al.(2021)Wang, Di~Renzo, Williams, Urzay, and
  Iaccarino]{wang2021progress}
J~Wang, M~Di~Renzo, C~Williams, J~Urzay, and G~Iaccarino.
\newblock Progress on laser ignition simulations of a ch4/o2 subscale rocket
  combustor using a multi-gpu task-based solver.
\newblock \emph{Center for Turbulence Research Annual Research Briefs}, pages
  129--142, 2021.

\bibitem[Passiatore et~al.(2024)Passiatore, Wang, Rossinelli, Di~Renzo, and
  Iaccarino]{passiatore2024computational}
Donatella Passiatore, Jonathan~M Wang, Diego Rossinelli, Mario Di~Renzo, and
  Gianluca Iaccarino.
\newblock Computational study of laser-induced modes of ignition in a coflow
  combustor.
\newblock \emph{Flow, Turbulence and Combustion}, pages 1--25, 2024.

\bibitem[Zahtila et~al.(2025)Zahtila, Cutforth, Brouzet, Passiatore,
  Rossinelli, and Iaccarino]{zahtila2025bi}
Tony Zahtila, Murray Cutforth, Davy~J Brouzet, Donatella Passiatore, Diego
  Rossinelli, and Gianluca Iaccarino.
\newblock Bi-fidelity data ensembles of a rocket ignition system with
  stochastic interpolative decomposition.
\newblock In \emph{AIAA SCITECH 2025 Forum}, page 2300, 2025.

\bibitem[Smagorinsky(1963)]{smagorinsky1963general}
Joseph Smagorinsky.
\newblock General circulation experiments with the primitive equations: I. the
  basic experiment.
\newblock \emph{Monthly weather review}, 91\penalty0 (3):\penalty0 99--164,
  1963.

\bibitem[Strelau et~al.(2023)Strelau, Frederick, Senior, Gejji, and
  Slabaugh]{strelau2023modes}
Ryan Strelau, Mark Frederick, Will~C Senior, Rohan Gejji, and Carson~D
  Slabaugh.
\newblock Modes of laser spark ignition of a model rocket combustor.
\newblock In \emph{AIAA SCITECH 2023 Forum}, page 2377, 2023.

\end{thebibliography}

\end{document}